# Improve Object Detection by Data Enhancement based on Generative Adversarial Nets

Wei Jiang，Na Ying

# ABSTRACT

The accuracy of the object detection model depends on whether the anchor boxes effectively trained. Because of the small number of GT boxes or object target is invariant in the training phase, cannot effectively train anchor boxes. Improving detection accuracy by extending the dataset is an effective way. We propose a data enhancement method based on the foreground-background separation model. While this model uses a binary image of object target random perturb original dataset image. Perturbation methods include changing the color channel of the object, adding salt noise to the object, and enhancing contrast. The main contribution of this paper is to propose a data enhancement method based on GAN and improve detection accuracy of DSSD. Results are shown on both PASCAL VOC2007 and PASCAL VOC2012 dataset. Our model with 321x321 input achieves 78.7% mAP on the VOC2007 test, 76.6% mAP on the VOC2012 test.

**Keywords**: Object Detection; GAN; Data enhancement

# 1. INTRODUCTION

Object detection includes positioning, identification and classification tasks. Object detection initially generates Prior Boxes by using the sliding-window method, RCNN[1] replace the sliding-window method with the selective-search method, it greatly reduces the number of Prior Box. Unlike, YOLO[2] and SSD[3] adopted a sliding-window-like method and multiplexed convolutional feature, synchronize processing positioning and classification tasks. Compared with FastRCNN[4] and FasterRCNN[5], such state-of-art object detection models like YOLO and SSD greatly increase the detection rate.

ResNet[6] proves that context information is beneficial to object classification. Similarly, FPN[7] combines high-level and low-level context information to improve object detection accuracy. These methods prove that multi-scale feature maps are beneficial to improve the detection accuracy, especially the detection accuracy of small object. However, combining high-level and low-level context information also increases computational complexity and network depth. Such methods lead to decrease the detection rate. If we continue to improve the accuracy of target detection with multi-scale feature maps, it will inevitably increase network depth and complexity.

Image translation model reflects the mapping relationship between the input image and the output image. Benefited from the rapid development of CNN and GAN[8], we could implement an image translation model simply and effectively. Based on U-NET[9], IMG2IMG[10] implements image domain mapping from street label images to real street images, object edge images to real images. CycleGAN[11] believes that the stable translate relationship cannot be formed by a pair of GAN networks, so it designed two pairs of GAN network structure based on ResNet to implement the local image translation. Combining the image translation model with the object detection model has recently become a trend. MaskRCNN[12] combined with FCN[13] proved that semantic segmentation helps to improve detection accuracy.

The main contribution of this paper is to propose a data enhancement method based on foreground-background separation model. Without changing the DSSD[14] main network, we added GAN network to the training process of DSSD. By not increasing the complexity of the original DSSD model, the GAN model is used to assist the DSSD network model training process. The test results on the PASCAL dataset indicate that this method did not increase the detection rate but improved the object detection accuracy.

## 2. RELATED WORK

The majority of object detection methods, including RCNN, FastRCNN, FasterRCNN, YOLO, and SSD, provides multiple ideas for improving object detection accuracy or object detection rate. RCNN adopted selective-search method instead of the sliding window method to improve detection rate. Inspired by SPPNET[15], FastRCNN and uses the same feature layer for position and classification to avoid duplicate counting. FasterRCNN replaces the selective-search method with RPN layer, which improves detection accuracy and achieves state-of-art detection model. Unlike RCNN-like model, YOLO treats the object detection problem as a regression problem and synchronizes the object positioning and classification tasks.

Further, SSD improves detection accuracy by using multi-scales convolutional feature layer with VGG[16] network. Inspired by the FPN, DSSD introduces context information in the SSD extra feature layer. It enables low-level features to share high-level features, further improving detection accuracy. However, use more multi-scale layers are bound to weaken the advantage of DSSD in detection rate.

GAN is a unique network structure that captures the distribution of potential data. The GAN network is optimized by minimizing G network loss and maximizing D network loss. The design of true and false loss based on game theory makes it possible to deal with different types of tasks. GAN has been widely used in the image-to-image translation field. IMG2IMG, CycleGAN established parameterized image mapping models on specific datasets. These methods prove the superiority of GAN in image translation task.

Based on FasterRCNN, MaskRCNN combines FCN to optimize the object detection model, it synchronizes semantic segmentation and object detection tasks. Similarly, PAD[17] uses semantic segmentation feature as network weight to optimize object detection. However, how to use the segmentation model to optimize YOLO and DSSD has not been effectively proved.

# 3. METHOD

DSSD improves the accuracy of object detection by adding more context feature. It is bound to decrease the detection rate. To improve detection accuracy and ensuring the detection rate. We propose a data enhancement method based on the foreground-background separation model. The overall network structure of the model is shown in Figure 3-1. The overall network can be divided into two parts, which mean two different training phases. The first phase achieves foreground-background model and pre-training object detection. The second phase uses data enhancement to assist object detection training. Next, we will discuss how to optimize the DSSD network training process.

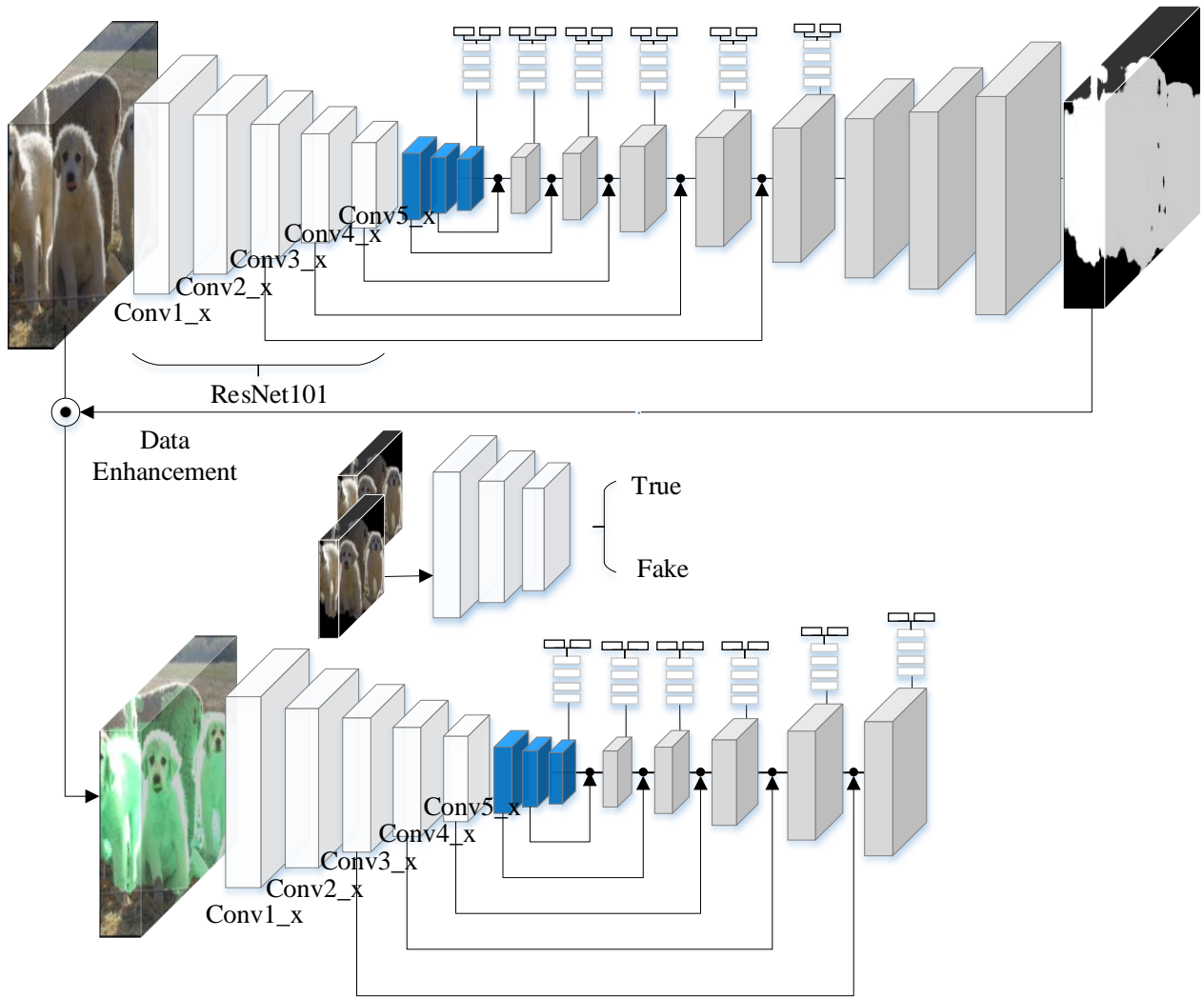

Figure 3-1 Network Arch

## 3.1 BASIC THEORY

DSSD is based on ResNet101, similar to SSD, DSSD adds several convolution layers at the end of the network, and add multiple DM deconvolution modules after the convolution layer to achieve multiplexing multi-scales convolution information. At the same time, the PM prediction module to predict the bounding box score and offset. Finally, using the NMS method to process PM prediction results.

GAN consists of two parts, including network G and network D. The network G generates fake image to confuse network D. The network D discriminates the fake image and the real image. Finally, the network G captures the real data distribution. The GAN model loss function is shown as follows:

$$L_{GAN}(G,D) = \min_G \max_D (E_{x \sim pdata(y)} \log(D(y)) + E_{x \sim pdata(x)} \log(1-D(x))) \qquad (3\text{-}1)$$

In the formula, $\min_G$ means to minimize the loss of network G, $\max_D$ means to maximize the loss of network D. $y \sim pdata(y)$ means input image data distribution, $x \sim pdata(x)$ means fake image data distribution. $E$ is the error mean.

## 3.2 ADJUST DSSD NETWORK STRUCTURE

For the broader use of the model, the modification for ResNet101 by DSSD is not adopted. We retain the main network structure of ResNet101 and selects conv3_x, conv4_x, conv5_x, and additional three feature layers. In the case of the 321x321 input size, the output size of each layer is shown in Table 3-1:

Table 3-1 feature map size of DSSD

| ResNet101 | conv3_x | conv4_x | conv5_x | conv6_x | conv7_x | conv8_x |
|---|---|---|---|---|---|---|
| Resolution | 41x41 | 21x21 | 11x11 | 6x6 | 3x3 | 2x2 |

To combine the GAN with the object detection network in the training process, we replace the DM deconvolution module multiplication method by the low-level feature and the high-level sign addition method. And we use the best performing PM-C prediction module for bounding box scores and offset predictions.

## 3.3 FOREGROUND BACKGROUND SEPARATION MODEL

Traditional image translation network such as IMG2IMG and CycleGAN have advantages in three-dimensional image domain mapping, but cannot effectively discriminate real or fake one-dimensional binarized foreground image. The reason is that binarized foreground image has a large amount of background redundancy information and interferes with the network D. To achieve an accurate mapping between input and output images, we propose the following adjustments.

(1) Similar to the traditional image translation model, we establish a specific input-output pair dataset. Based on this dataset, we train a foreground-background separation translation model. The mapping relationship is shown in Figure 3-2. (a1) and (b1) are original images $I$, (a2) and (b2) are binarized foreground image $I_{MASK}$.

(a1)      (a2)      (b1)      (b2)

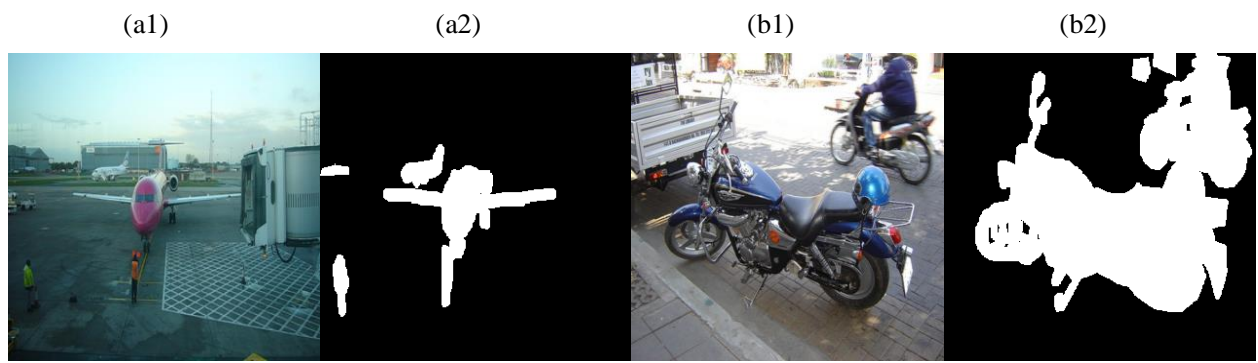

Figure 3-2 Dataset mapping relationship

(2) In addition to the GAN original loss, we define a regular loss to achieve the mapping from the three-dimensional image to the one-dimensional image. Because of the background information of binarized foreground images interfere with network D. Inspired by Perceptual GAN, we add perceptual loss as a regular loss in the foreground-background separation model. The perceptual loss is shown in Equation 3-2:

$$L_P(G) = \sum_{i=1}^{j} \mathrm{E}_{x,y \sim p_{data}(x,y)}\left[\|fm_i(x) - fm_i(y)\|\right] \qquad (3\text{-}2)$$

Where $[fm_1, fm_2, \cdots, fm_i]$ is the middle layer feature of network D. $fm_i(x)$ means the $ith$ layer feature obtained by discriminating the fake images, and $fm_i(y)$ means the $ith$ layer feature obtained by discriminating the real images. Then we calculate the L2-norm loss between $fm_i(x)$ and $fm_i(y)$.

Due to the loss of a large amount of information in the binarized foreground image, it is difficult for network D to discriminates true and false images. In this paper, the image $I$ is multiplied by the image $I_{MASK}$ as the input of network D. In this way, the background information of the image $I$ is filtered out, and the information of the object is added to the binarized foreground image. As shown in Figure 3-3, from left to right are image $I$, image $I_{MASK}$, and image $I_{CROP}$.

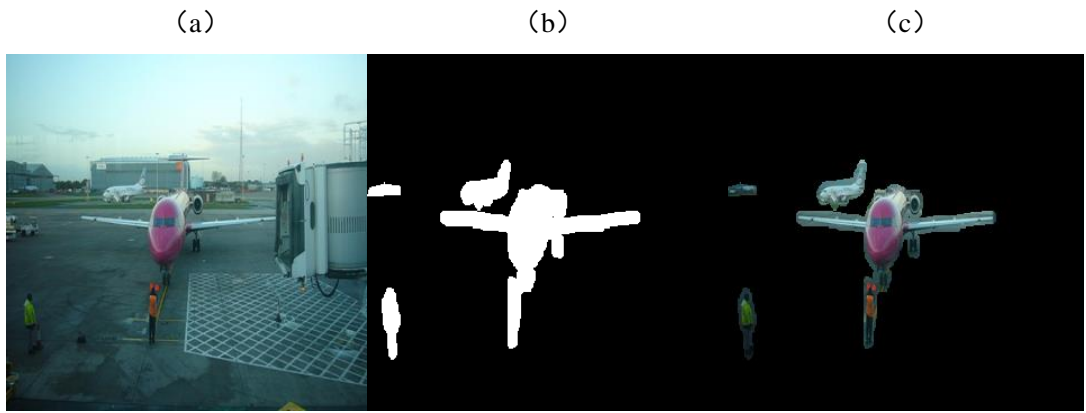

Figure 3-3 The result of image multiplication

The result of the foreground-background separation model is shown in Figure 3-4.

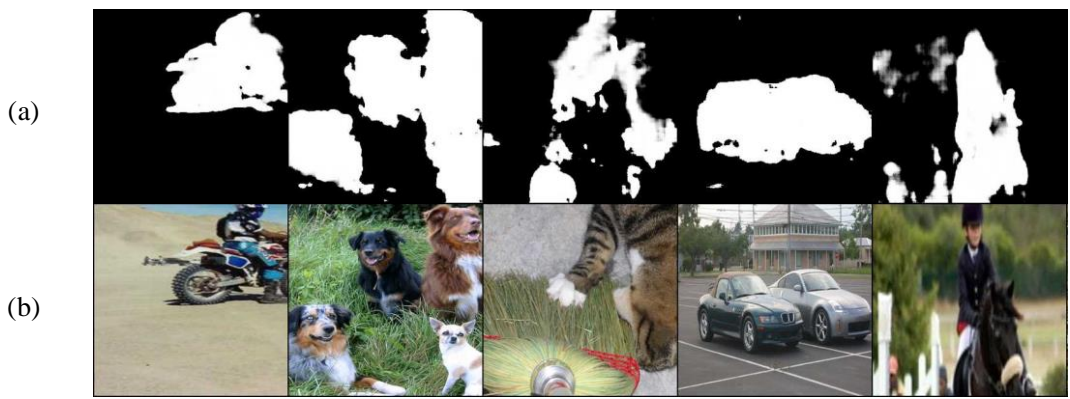

Figure 3-4 The result of the foreground-background separation model

(a) is the output of the foreground-background separation model, and (b) is the original input image

of the foreground-background separation model.

Contrary to the traditional FCN semantic separation network, the foreground-background separation model separate the foreground and background of the image, instead of separating the different categories. Based on GAN, the loss function and the convolutional network takes up less memory. The network structure is shown in Figure 3-1.

## 3.4 DATA ENHANCEMENT

Whether it is FasterRCNN or DSSD, data enhancement has proven effective for improving detection accuracy. The data enhancement in FasterRCNN randomly crop the original image and add random photometric distortion and random flip. On this basis, DSSD and SSD use random enhancement technique to enhance small-object detection accuracy.

The detection mAP of the DSSD model without data enhancement is shown in Table 3-2.

Table 3-2. The mAP of DSSD without data enhancement

| Method  | Datasets | Network | mAP   |
|---------|----------|---------|-------|
| DSSD321 | 07+12    | Res101  | 65.4% |

However, the common point of the traditional data enhancement based on the full image, resulting in many images different from the original images in the dataset. This section describes the data enhancement method based on the GAN. This method enhances data by distinguishing the foreground and background of the image. It is beneficial to classifies foreground and background for object detection model.

(1) During the second training process, this data enhancement method is performed with a certain probability. For the same batch of eight pictures, the probability of data enhancement is 0.5. And randomly select the top N pictures in the range of (0, 8) for data enhancement.

The part of the foreground-background separation model structure is shown in Figure 3-5.

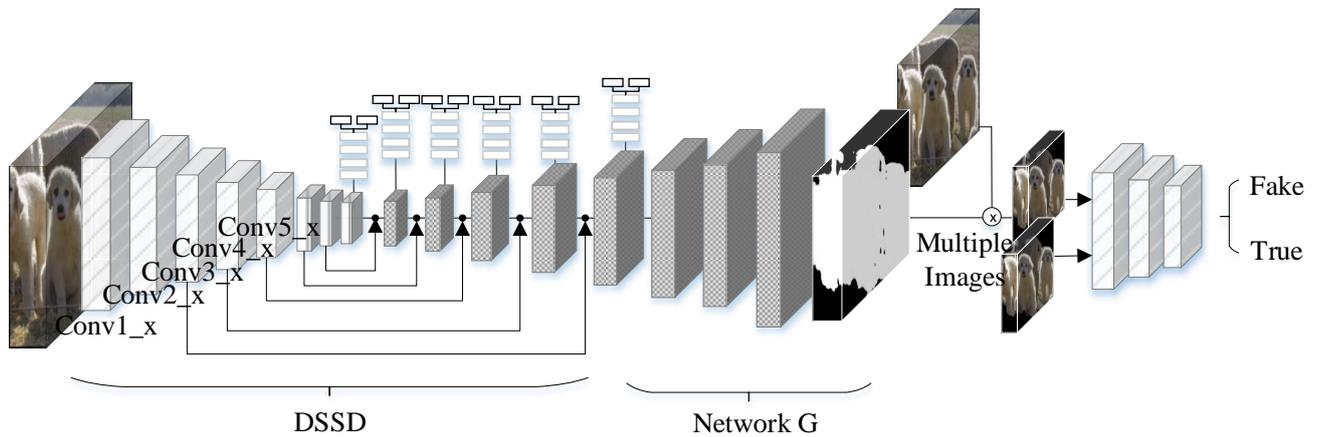

Figure 3-5 Foreground background separation model

Figure 3-5 Network layer definition is consistent with Figure 3-1. The input of this model in Figure 3-5 comes from the DSSD network. The network G generate the binarized foreground image $I_{MASK}$. Image $I_{MASK}$ and the image $I$ is multiplied to get a true or false foreground image $I_{CROP}$. Then we use image $I_{CROP}$ as the input of network D.

(2) The binarized image is generated from the foreground-background separation model in

Figure 3-5. Based on the binarized image $I_{MASK}$, the object in the original image $I$ is randomly disturbed, and the background of the original image $I$ is retained. The random perturbation modes include changing the object color channel, adding noise to the object, and enhancing the color contrast. The result of data enhancement is shown in Figure 3-6.

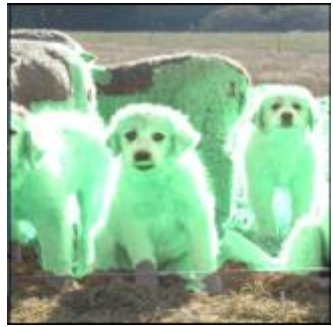

Figure 3-6 Data enhancement example

During the second training process, the image after the data enhancement is reused as the input of the object detection model, as shown in Figure 3-7:

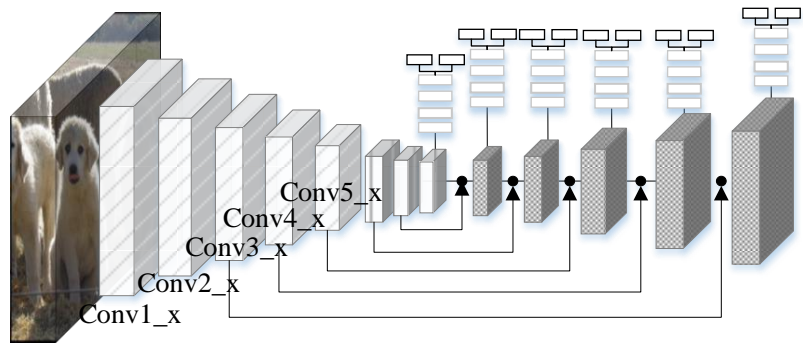

Figure 3-7 Data Enhancement Training Network Structure

Figure 3-7 shows the detailed of the network structure at the bottom of Figure 3-1. The network layer definition is consistent with Figure 3-1. During the data enhancement training phase, network shares the network weight with the top network structure of Figure 3-1. Therefore, the object-based data enhancement method proposed in this paper does not change the network complexity in the detection stage, just increased the complexity of the network in the training stage.

The detection results of the data enhancement methods proposed in this paper are shown in Table 4-2 and Table 4-3.

# 4 EXPERIMENTS

## 4.1 TRAINING DETAILS

The key to the SSD method is the Prior Box, DSSD adds a new Prior Box scale of 1.6 to each feature layer. This article fine-tunes the Prior Box proportional scale design in DSSD. As shown in Table 4-1:

Table 4-1 Prior Box scale design

| ResNet101 | conv3_x | conv4_x | conv5_x | conv6_x | conv7_x | conv8_x |
| --- | --- | --- | --- | --- | --- | --- |
| Resolution | 41 x 41 | 21x 21 | 11x 11 | 6x 6 | 3x 3 | 2x 2 |
| Aspectratios | 2 | 2,3,1.6 | 2,3,1.6 | 2,3,1.6 | 2 | 2 |

In the first phase, we directly train the DSSD network model. And in the second phase, we apply data-enhancement method based on GAN to assist DSSD.

Whenever the first or second phases of the model training phase, we retain the data enhancement methods adopted by FasterRCNN and DSSD. The following experiments were performed in the GPU 1050ti environment.

## 4.2 PASCAL VOC2007

The model was trained based on PASCAL VOC2007 and PASCAL VOC2012 trainval datasets. In the first training phase, we used a batch size of 8 for the model with 321x321 inputs, started the learning rate at 1e-3 for the first 120 epochs. Then, we decreased it to 1e-4 during 120 epochs to 160 epochs. Finally, we decreased it to 1e-5 during 160 epochs to 200 epochs.

During the second training phase, similar to DSSD, the DSSD parameter does not change during the foreground-background separation model training phase. Base on the input-output pair dataset of Figure 3-2, we used a batch size of 8 for the model 321x321 inputs and set the learning rate at 1e-3 during 200 epochs to 220 epochs, a foreground-background separation model is obtained.

After training the foreground-background separation model, we modified the original image $I$ by the output of the foreground-background separation model in the data enhancement modes as shown in Section 3.3. Then we reuse those modified images as the input for the object detection network. Table 4-2 shows our results on PASCAL VOC2007 test datasets.

Table 4-2 PASCAL VOC2007 test mAP

| Method | Datasets | Network | mAP |
| --- | --- | --- | --- |
| SSD300 | 07+12 | VGG | 77.5% |
| SSD321 | 07+12 | Resnet101 | 77.1% |
| DSSD321 | 07+12 | Resnet101 | 78.6% |
| OurModel | 07+12 | Resnet101 | 78.7% |

The full detection results are shown in Table-a and figure-a in the appendix.

Compare with the design of SSD 321+PM-C+DM(Eltw-sum) in DSSD. Our model achieved 78.7% mAP better than DSSD 78.4% mAP in the same situation, and the best result of DSSD 78.6% mAP.

## 4.3 PASCAL VOC2012

For the PASCAL VOC 2012 task, during the training phase, we load a pre-trained PASCAL VOC2007 model and foreground-background separation model. We use 07++12 datasets consisting of PASCAL VOC2007 trainval, PASCAL VOC2012 trainval, and PASCAL VOC2007 test. We used a batch size of 8 for the model with 321x321 input and started the learning rate at 1e-3 during 200 epochs to 240 epochs. Table 4-3 shows our results on PASCAL VOC2012 test datasets.

Table 4-3 PASCAL VOC2012 test mAP

| Method | Datasets | Network | mAP |
| --- | --- | --- | --- |
| SSD300 | 07++12 | VGG | 75.8% |
| SSD321 | 07++12 | Resnet101 | 75.4% |
| DSSD321 | 07++12 | Resnet101 | 76.3% |
| OurModel | 07++12 | Resnet101 | 76.6% |

The full detection results are shown in Table-b and Figure-b in the appendix.

# 5 SUMMARY

Based on the DSSD network, we propose a method that does not increase the complexity of the original network structure to improve the DSSD object detection accuracy. The method proposed in this paper has been proved to be effective in the PASCAL VOC dataset. Because this method is independent of the object detection method, it is also applicable to other object detection network models.

# REFERENCES

[1]. Girshick R, Donahue J, Darrell T, et al. Rich Feature Hierarchies for Accurate Object Detection and Semantic Segmentation[C]// IEEE Conference on Computer Vision and Pattern Recognition. IEEE Computer Society, 2014:580-587.

[2]. Redmon J, Divvala S, Girshick R, et al. You only look once: Unified, real-time object detection[C]//Proceedings of the IEEE conference on computer vision and pattern recognition. 2016: 779-788.

[3]. Liu W, Anguelov D, Erhan D, et al. Ssd: Single shot multibox detector[C]//European conference on computer vision. Springer, Cham, 2016: 21-37.

[4]. Girshick R. Fast r-cnn[C]//Proceedings of the IEEE international conference on computer vision. 2015: 1440-1448.

[5]. Ren S, He K, Girshick R, et al. Faster R-CNN: Towards Real-Time Object Detection with Region Proposal Networks[J]. IEEE Transactions on Pattern Analysis & Machine Intelligence, 2017, 39(6):1137-1149.

[6]. He K, Zhang X, Ren S, et al. Deep residual learning for image recognition[C]//Proceedings of the IEEE conference on computer vision and pattern recognition. 2016: 770-778.

[7]. Lin T Y, Dollár P, Girshick R, et al. Feature pyramid networks for object detection[C]//CVPR. 2017, 1(2): 4.

[8]. Goodfellow I, Pouget-Abadie J, Mirza M, et al. Generative adversarial nets[C]//Advances in neural information processing systems. 2014: 2672-2680.

[9]. Ronneberger O, Fischer P, Brox T. U-net: Convolutional networks for biomedical image segmentation[C]//International Conference on Medical Image Computing and Computer-Assisted Intervention. Springer, Cham, 2015: 234-241.

[10]. Isola P, Zhu J Y, Zhou T, et al. Image-to-image translation with conditional adversarial networks[J]. arXiv preprint arXiv:1611.07004, 2016.

[11]. Zhu J Y, Park T, Isola P, et al. Unpaired image-to-image translation using cycle-consistent adversarial networks[J]. arXiv preprint arXiv:1703.10593, 2017.

[12]. He K, Gkioxari G, Dollár P, et al. Mask r-cnn[C]//Computer Vision (ICCV), 2017 IEEE International Conference on. IEEE, 2017: 2980-2988.

[13]. Long J, Shelhamer E, Darrell T. Fully convolutional networks for semantic segmentation[C]//Proceedings of the IEEE conference on computer vision and pattern recognition. 2015: 3431-3440.

[14]. Fu C Y, Liu W, Ranga A, et al. DSSD: Deconvolutional single shot detector[J]. arXiv preprint arXiv:1701.06659, 2017.

[15]. He K, Zhang X, Ren S, et al. Spatial pyramid pooling in deep convolutional networks for visual recognition[C]//european conference on computer vision. Springer, Cham, 2014: 346-361.

[16]. Simonyan K, Zisserman A. Very deep convolutional networks for large-scale image recognition[J]. arXiv preprint arXiv:1409.1556, 2014.

[17]. Zhao X, Liang S, Wei Y. Pseudo Mask Augmented Object Detection[J]. arXiv preprint arXiv:1803.05858, 2018.

[18]. Wang C, Xu C, Wang C, et al. Perceptual Adversarial Networks for Image-to-Image Transformation[J]. arXiv preprint arXiv:1706.09138, 2017.

# APPENDIX

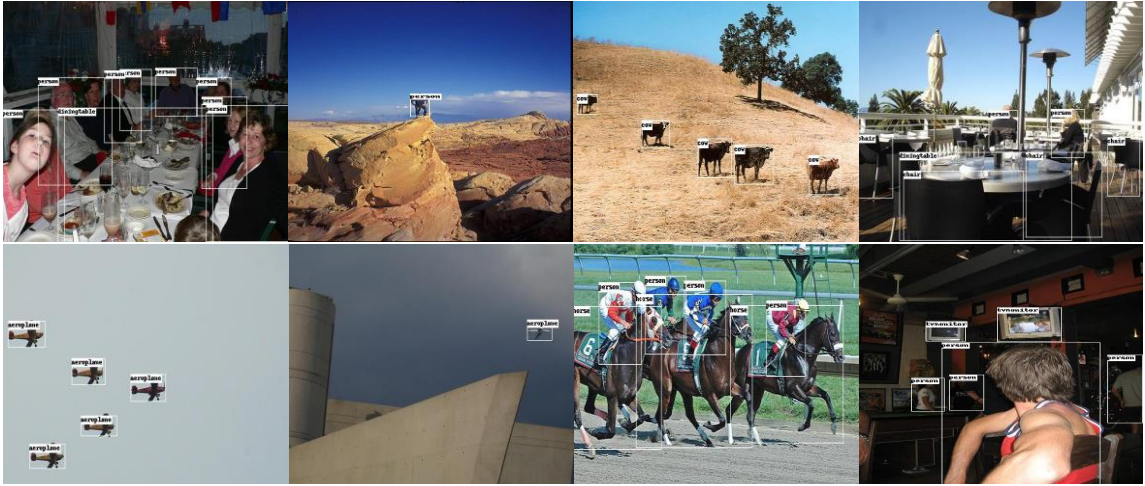

Figure a. PASCAL VOC2007 detection results

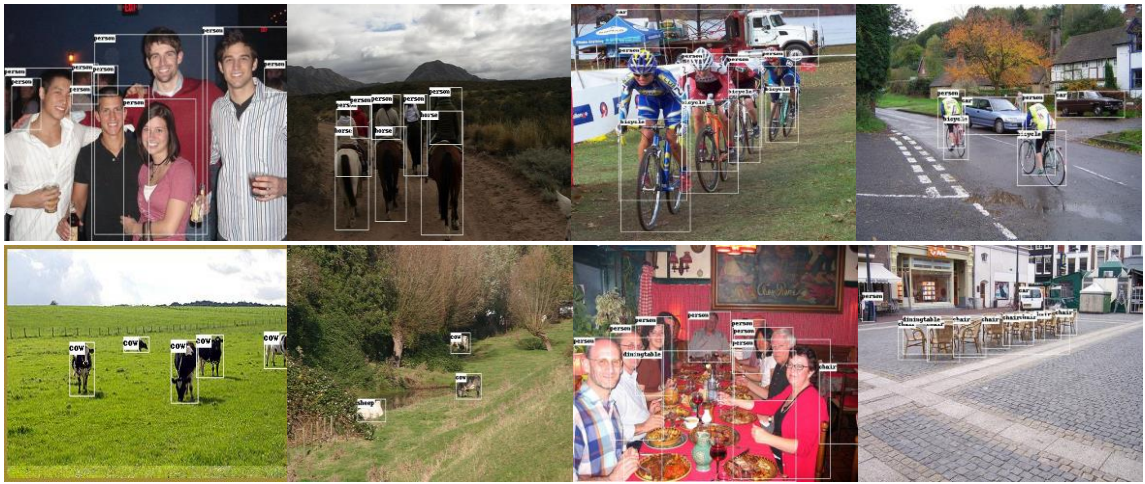

Figure b. PASCAL VOC2012 detection results

Table a PASCAL VOC2007 test mAP

| Method | SSD300 | SSD321 | DSSD321 | OurModel |
| --- | --- | --- | --- | --- |
| Datasets | 07+12 | 07+12 | 07+12 | 07+12 |
| Network | VGG | ResNet101 | ResNet101 | ResNet101 |
| mAP | 77.5 | 77.1 | 78.6 | 78.7 |
| Aero | 79.5 | 76.3 | 81.9 | 79.1 |
| Bike | 83.9 | 84.6 | 84.9 | 86.7 |
| Bird | 76.0 | 79.3 | 80.5 | 78.5 |
| Boat | 69.6 | 64.6 | 68.4 | 72.2 |
| Bottle | 50.5 | 47.2 | 53.9 | 54.8 |
| Bus | 87.0 | 85.4 | 85.6 | 84.6 |
| Car | 85.7 | 84.0 | 86.2 | 86.6 |
| Cat | 88.1 | 88.8 | 88.9 | 87.1 |
| Chair | 60.3 | 60.1 | 61.1 | 64.9 |
| Cow | 81.5 | 82.6 | 83.5 | 82.1 |
| Table | 77.0 | 76.9 | 78.7 | 75.5 |
| Dog | 86.1 | 86.7 | 86.7 | 84.9 |
| Horse | 87.5 | 87.2 | 88.7 | 87.5 |
| Mbike | 83.97 | 85.4 | 86.7 | 86.4 |
| Person | 79.4 | 79.1 | 79.7 | 79.7 |
| Plant | 52.3 | 50.8 | 51.7 | 57.0 |
| Sheep | 77.9 | 77.2 | 78.0 | 79.9 |
| Sofa | 79.5 | 80.9 | 82.6 | 79.5 |
| Train | 87.6 | 87.3 | 87.2 | 86.5 |
| TV | 76.8 | 76.6 | 79.4 | 78.4 |

Table b PASCAL VOC2012 test mAP

| Method | SSD300 | SSD321 | DSSD321 | OurModel |
| --- | --- | --- | --- | --- |
| Datasets | 07++12 | 07++12 | 07++12 | 07++12 |
| Network | VGG | ResNet101 | ResNet101 | ResNet101 |
| mAP | 75.8 | 75.4 | 76.3 | 76.6 |
| Aero | 88.1 | 87.9 | 87.3 | 88.1 |
| Bike | 82.9 | 82.9 | 83.3 | 84.6 |
| Bird | 74.4 | 73.7 | 75.4 | 76.2 |
| Boat | 61.9 | 61.5 | 64.6 | 61.9 |
| Bottle | 47.6 | 45.3 | 46.8 | 51.3 |
| Bus | 82.7 | 81.4 | 82.7 | 83.7 |
| Car | 78.8 | 75.6 | 76.5 | 78.5 |
| Cat | 91.5 | 92.6 | 92.9 | 91.9 |
| Chair | 58.1 | 57.4 | 59.5 | 58.1 |
| Cow | 80.0 | 78.3 | 78.3 | 83.1 |
| Table | 64.1 | 65.0 | 64.3 | 64.9 |
| Dog | 89.4 | 90.8 | 91.5 | 91.3 |
| Horse | 85.7 | 86.8 | 88.6 | 87.5 |
| Mbike | 85.5 | 85.8 | 86.6 | 84.7 |
| Person | 82.6 | 81.5 | 82.1 | 82.6 |
| Plant | 50.2 | 50.3 | 53.3 | 52.9 |
| Sheep | 79.8 | 78.1 | 79.6 | 80.7 |
| Sofa | 73.6 | 75.3 | 75.7 | 70.3 |
| Train | 86.6 | 85.2 | 85.2 | 85.9 |
| TV | 72.1 | 72.5 | 73.9 | 75.1 |